# Analyse et structuration automatique des guides de bonnes pratiques cliniques : essai d'évaluation

Amanda Bouffier[1], Thierry Poibeau[1], Catherine Duclos[2]

[1] Laboratoire d'Informatique de Paris-Nord (LIPN),
UMR CNRS 7030 et Université Paris 13
{prenom.nom}@lipn.univ-paris13.fr

[2] Laboratoire d'Informatique Médicale et BioInformatique (EA 3969 - LIM&BIO),
UFR SMBH, Université Paris 13
catherine.duclos@avc.aphp.fr

**Communication appliquée**

**Résumé :**

Les guides de bonnes pratiques cliniques (GBPC) sont des textes constitués de recommandations valides dont le but est de diffuser des synthèses de résultats démontrés et de normaliser des conduites à tenir dans des situations cliniques données. L'adhésion des médecins à ces guides doit conduire à une médecine de qualité basée sur des preuves scientifiques. Cet article présente un outil appelé GemFrame, destiné à faciliter la consultation des guides en proposant de nouveaux modes d'accès sur support électronique. Pour ce faire, un travail d'analyse et de structuration des GBPC est nécessaire.

Nous présentons le système GemFrame, permettant cette structuration suite à une analyse semi-automatisée. Les GBPC étant des textes d'« incitation à l'action », ils sont principalement constitués de conditions et d'actions dépendant de ces conditions. Le système GemFrame vise à reconnaître automatiquement les segments « conditions » et les « segments incitation à l'action », puis à calculer la portée des conditions sur les actions. Nous présentons ici une évaluation détaillée sur plusieurs guides. Nous montrons d'abord l'intérêt de l'approche puis nous détaillons le processus d'évaluation fondé sur la comparaison des résultats obtenus automatiquement avec ceux obtenus manuellement (suite à l'élaboration de « référence »).

## 1 Introduction

Les guides de bonnes pratiques cliniques (GBPC) sont des textes constitués de recommandations valides dont le but est de diffuser des synthèses de résultats démontrés (obtenus à partir d'essais cliniques) et de normaliser des conduites à tenir dans des situations cliniques données. L'adhésion des médecins à ces guides doit conduire à une médecine de qualité basée sur des preuves scientifiques.

Bien que l'utilité de ces guides soit évidente, leur utilisation en pratique connaît un certain nombre d'obstacles. Parmi ces obstacles, il est possible d'identifier des facteurs intrinsèques liés au médecin et à sa capacité à connaître, adhérer et croire au guide [Cabana et al 1999], mais aussi des facteurs liés au texte lui-même. En effet, ces textes issus de consensus d'experts, présentent un niveau de complexité élevé (longueur, manque de standardisation, manque de clarté, construction complexe faisant intervenir des relations causales ou temporelles) [Patel et al 2001] [Elkin et al 2000]. Ces textes peuvent également présenter des ambiguïtés [Codish 2005], des imprécisions, ou ne pas traiter l'ensemble des situations possibles [Shiffman, Greene 1994], [Shiffman et al 1999],

L'amélioration de ces documents en en facilitant la compréhension, en en clarifiant leur contenu, en en contrôlant la cohérence semble pouvoir faciliter leur diffusion [Patel et al 2001] , [Shiffman 2004], [Shiffman 2003]. Des modèles de représentation des connaissances contenues dans les guides ainsi que des formats documentaires ont été proposés afin de structurer les guides, les rendre plus homogènes et plus facilement lisibles.

Cet article présente un outil appelé GemFrame, destiné à faciliter la structuration des guides par un processus semi-automatisé. Les GBPC étant des textes d'incitation à l'action, il sont principalement constitués de conditions et d'actions dépendant de ces conditions. Le système GemFrame vise à reconnaître automatiquement les segments « conditions » et les « segments incitation à l'action », puis à calculer la portée des conditions sur les actions. Plutôt que d'insister sur le mode de fonctionnement de l'outil qui a été précisé dans diverses publications [Bouffier et Poibeau, 2007 ; Bouffier, 2007], nous présentons ici une évaluation détaillée sur plusieurs guides. Nous montrons d'abord l'intérêt de l'approche puis nous détaillons le processus d'évaluation fondé sur la comparaison des résultats obtenus automatiquement avec ceux obtenus manuellement (suite à l'élaboration de « référence »).

Nous revenons dans un premier temps sur l'état de l'art afin de montrer les principales réalisations dans ce cadre. Nous présentons ensuite rapidement le fonctionnement de GemFrame avant d'en venir à l'évaluation du système, effectuée automatiquement à partir d'une comparaison avec un découpage manuel. Nous terminons en discutant les résultats obtenus et leur pertinence pour la tâche.

## 2  Vers une modélisation automatique des GBPC

Nous présentons dans cette partie différents formalismes permettant de rendre compte des GBPC, ainsi que les outils permettant une automatisation plus ou moins grande du processus de modélisation.

### 2.1  Intérêt de modéliser les GBPC

Des modèles et des formalismes de représentation de connaissances ont été proposés pour structurer ces guides et les informatiser. L'intérêt de l'informatisation des GBPC est qu'il est alors possible d'envisager un certain nombre de fonctionnalités simplifiant la gestion des étapes du cycle de vie du guide de bonne pratique. Ainsi, il peut être possible d'identifier l'évolution des connaissances entre deux versions d'un guide et d'envisager la maintenance automatisée de la base de connaissances [Shahar 2004], [Shalom 2005], [Shiffman 2000]. Il peut être aussi possible de vérifier la cohérence du guide avec une élaboration automatique des arbres de décision et une recherche automatique des situations manquantes [shiffman 1997]. Enfin l'informatisation peut permettre d'aider à l'écriture du guide et en faciliter l'intégration dans des systèmes d'aide à la décision .

L'intégration des guides de bonnes pratiques dans des systèmes d'aide à la décision médicale est un moyen d'apporter au médecin, dans son environnement de travail, un accès à la connaissance médicale. Diverses modalités d'intégrations peuvent être envisagées. Par exemple, dans le cadre du projet ASTI (Aide à la Stratégie Thérapeutique Informatisée), deux modes d'utilisation des connaissances issues des guides de bonnes pratiques ont été envisagés : le premier est un mode guidé qui aide le médecin à naviguer dans l'arborescence du guide en fonction des caractéristiques de son patient pour aboutir à une décision (par exemple le traitement à prescrire) [Séroussi 2001]. Le mode critique constitue, lui, le deuxième mode d'utilisation des connaissances issues du GBPC. Il fonctionne en feedback d'une décision du médecin : le système confronte la décision thérapeutique du médecin à celle que le système trouve utilisant les données d'historique thérapeutique, les données cliniques du patient et les règles de décisions de la base de connaissance et construit une critique de la prescription du médecin [Ebrahiminia 2006].

### 2.2  Quelles connaissances dans les différents modèles de représentation des connaissances des GBPC ?

Les guides de bonnes pratiques cliniques peuvent être vus comme un ordonnancement de décision et d'actions. Les formalismes de représentation développés visent à identifier ces éléments et proposer des modalités d'ordonnancement dans le temps [De Clerc 2004]. On peut distinguer deux approches : dans la première, les recommandations sont des décisions dépendant des états du patient et des modalités de soins (*state-based modeling*) dans la deuxième, les recommandations sont des séquences d'actions et de décisions mises en oeuvre pour atteindre un certain objectif (*plan-based modeling*) .

L'approche EON par exemple propose une modélisation par les états du patients sous la forme de scénarii dans lesquels les actions et décisions vont entraîner un changement de l'état du patient au cours du temps.

Le formalisme GLIF (*Guideline Interchange Format*) propose une modélisation par la planification des soins et permet de modéliser les guides sous la forme de diagramme algorithmique en décrivant des étapes (*guideline steps*) ordonnées dans le temps. Ces étapes sont décrites sous la forme de conditions, d'actions, de décisions, de modalités d'exécution des étapes (simultanéité, synchronisation).

### 2.3 Moyens mis en oeuvre pour aider à la structuration des guides

La plupart de ces formalismes sont assez complexes et la formalisation manuelle des guides selon ces modèles est source de difficulté (compréhension du modèle) et de variabilité entre les individus. Aussi pour supporter cette formalisation des connaissances dans les GBPC, des outils permettant le balisage du texte ont été créés, reposant sur des modèles documentaires (comme GEM – *Guideline Element Model* [Shiffman 2000])

La structuration dans certains de ces éditeurs est manuelle, l'utilisateur choisit les composants à structurer au moyen d'une interface graphique (GEM Cutter, Guide-X/Stepper,GMT/DELT/A,DeGeL/URUZ). Cette approche manuelle conduit à une grande variabilité dans la structuration car l'interprétation des guides peut être différente d'un individu à l'autre.

D'autres outils proposent un traitement du document en préalable à la structuration par repérage de marqueurs caractéristiques du texte (opérateurs déontiques dans G-DEE [Georg 2005]) ou par apprentissage [Kaiser 2007]. L'approche systématique présente l'intérêt de la reproductibilité. C'est dans cette perspective qu'a été développé le système GemFrame.

## 3  Modélisation des guides de bonnes pratiques cliniques : le système GemFrame

Nous présentons ici le système GemFrame, dédié au repérage de séquences conditions-recommandations au sein des GBPC. L'originalité de ce système est de calculer la portée des segments conditionnels, ce qui constitue une réelle valeur ajoutée par rapport aux systèmes similaires déjà existants ([Georg 2005]).

### 3.1  Un Traitement en deux étapes

Le traitement que nous proposons est constitué de deux étapes principales : 1) repérage des segments élémentaires « condition » et « recommandations » ; 2) calcul de la portée des segments conditionnels (ce qui revient à déterminer, pour une

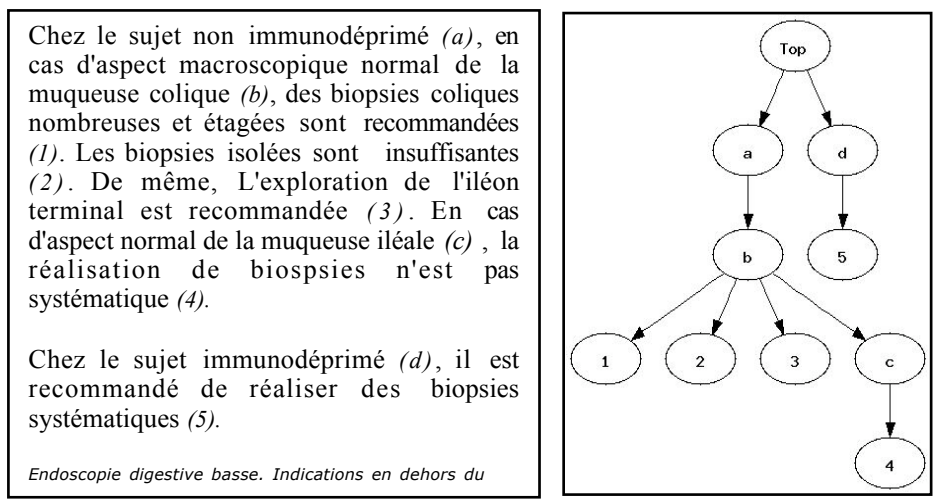

Chez le sujet non immunodéprimé *(a)*, en cas d'aspect macroscopique normal de la muqueuse colique *(b)*, des biopsies coliques nombreuses et étagées sont recommandées *(1)*. Les biopsies isolées sont insuffisantes *(2)*. De même, L'exploration de l'iléon terminal est recommandée *(3)*. En cas d'aspect normal de la muqueuse iléale *(c)*, la réalisation de biospsies n'est pas systématique *(4)*.

Chez le sujet immunodéprimé *(d)*, il est recommandé de réaliser des biopsies systématiques *(5)*.

*Endoscopie digestive basse. Indications en dehors du*

**Figure 1 :** Du texte à un arbre représentant la portée des segments conditionnels

recommandation donnée, sous quelle condition celle-ci est vraie). Au final, la structure générée automatiquement est un arbre XML dont les feuilles sont des segments « recommandations » et les noeuds des segments conditionnels. Tous les descendants d'un noeud condition $c$ (c'est-à-dire une liste de recommandations ou de conditions dans le cas de structures imbriquées) sont sous la portée de $c$ (autrement dit, d'un point de vue véridictionnel, toutes les recommandations vraies uniquement sous cette condition et non de façon générale). La figure 1 montre un exemple d'arbre produit par le système.

L'ensemble du traitement est réalisé automatiquement. Le résultat doit néanmoins être validé de manière interactive par un expert : notre système constitue donc une « aide à la modélisation ». Dans ce qui suit, nous présentons les deux étapes de traitement et de manière plus détaillée les règles permettant de calculer la portée des segments conditionnels.

### 3.2  Segmentation élémentaire

La première étape consiste à découper le texte en segments élémentaires « condition » et « recommandation ». Nous repérons dans un premier temps les *marques linguistiques* exprimant la condition et la recommandation. Nous utilisons pour cela une base de règles qui exploitent des classes de *marqueurs linguistiques* (tels que des verbes comme *recommander*, *conseiller* ou des adjectifs tels que *nécessaire*, *important*, *etc.*) acquis semi-automatiquement à partir de régularités repérées en corpus.

Dans un second temps, il s'agit de délimiter ces segments élémentaires. Afin de simplifier les traitements, nous avons fait le choix de ne pas segmenter en deçà du niveau de la phrase ou du niveau de la proposition, lorsque celle-ci est délimitée graphiquement4. C'est pourquoi cette segmentation s'appuie essentiellement sur des délimiteurs physiques. Un ensemble de règles étend les segments jusqu'à la présence de délimiteurs retenus comme pertinents (début paragraphe, fin de phrase, fin d'une énumération *etc*.).

À l'issue de cette étape, les segments élémentaires conditions et actions ont été repérés. Le premier cadre de la figure 1 montre un exemple de ce qui est produit en sortie de cette étape (les segments sont numérotés).

### 3.3 Calcul de la portée des segments conditionnels.

Une fois les segments élémentaires repérés, il faut repérer les cadres introduits par les segments « condition ». Cette étape repose en grande partie sur la position dans le texte des segments conditions repérés à l'étape précédente. Ceux-ci donnent lieu à une segmentation par défaut qui peut être contredite en présence de marques de rupture ou de continuité.

#### 3.3.1 Indices typo-dispositionnels pour le calcul des cadres

Une étude manuelle de plusieurs guides [Bouffier, 2006] a montré que la position des segments condition joue un rôle déterminant pour la segmentation en cadres Nous distinguons deux cas :

– L'introducteur est une expression non intégrée syntaxiquement (ce cas inclut les expressions détachées en début de phrase, mais aussi les titres et les amorces d'énumération).
– L'introducteur est une expression intégrée syntaxiquement à la phrase.

Les introducteurs en position détachée ont une propension à avoir une portée étendue (au-delà de la phrase courante) alors que les introducteurs en position intégrée ont une propension à avoir une portée minimale. Plus précisément, l'étude déjà citée a montré que la majorité des expressions détachées introduisent un cadre qui se ferme à la fin du paragraphe courant alors que la majorité des expressions intégrées introduisent un cadre qui se ferme à la fin de leur propre phrase. Ceci permet de fonder l'analyse sur une segmentation par défaut.

1. Les titres ont une portée qui s'étend jusqu'au titre suivant de même niveau ;
2. Les amorces d'énumération ont une portée qui couvre tous les éléments de l'énumération ;
3. Dans le cas des expressions détachées : la segmentation par défaut est égale au paragraphe ;
4. Dans le cas des expressions intégrées : la segmentation par défaut est égale à la phrase.

Pour les cas 3 et 4, suivant les GBPC considérés, la segmentation par défaut couvre 50 à 80 % des cas. Elle est toutefois remise en cause quand la présence d'autres indices indique une continuité entre différents paragraphes ou, à l'inverse, une rupture au sein du paragraphe.

### 3.3.2 Cas de remise en cause de la segmentation par défaut

Nous présentons dans cette section, de manière non systématique, quelques cas où la segmentation par défaut est remise en cause du fait de la présence d'indices de rupture ou de continuité au sein du texte.

***Redondance entre le titre et le premier introducteur de condition (exception à la règle 3).*** Nous avons observé des cas relativement fréquents où le titre est en partie redondant avec le premier introducteur. Bien que celui-ci soit en position détachée, sa portée dépasse le paragraphe et se confond avec celle du titre. Il s'agit donc d'une exception à la règle 3.

***Présence de marques de rupture discursive (exception à la règle 3).*** Certaines marques discursives sont à même de signaler la fermeture prématurée d'un cadre avant la fin de paragraphe, alors même que ce cadre est introduit par une expression détachée (exception à la règle 3). Ces marques peuvent marquer un contraste (à travers des formes lexicales comme *cependant, en revanche…*) ou une justification (*en effet, en fait…*)[1].

***Présence de marques de relations anaphoriques.*** Certaines relations anaphoriques sont des indices privilégiés pour signaler la continuation d'un cadre après la fin de la phrase, alors même que son introducteur est une expression détachée.

L'ensemble de ces éléments a été implémenté à travers un système multi-agent permettant de gérer les multiples contraintes intervenant dans le processus de structuration des GBPC. Le système ainsi constitué reçoit en entrée un GBPC et fournit en sortie un GBPC structuré d'après le modèle GEM. Il est alors possible de comparer ce processus avec une structuration manuelle des GBPC afin d'en évaluer les performances.

## 4   Évaluation

Les outils aidant à la formalisation des GBPC doivent être capables de proposer une analyse du guide proche de celle d'un expert. Pour appréhender leur performance, il convient de comparer leurs résultats à ceux d'un groupe d'experts sur des guides n'ayant pas servi à leur développement.

---

[1] Bien que ces marques soient généralement considérées comme des marques de cohésion, elles nous servent ici à identifier la fin du cadre, dans la mesure où la suite du texte correspond à une séquence étiquetée « Justification » suivant la DTD GEM.

### 4.1 Principes généraux sur l'évaluation

Plusieurs approches peuvent être envisagées. La première approche consiste à présenter à l'expert le texte balisé et de lui demander si ce balisage est correct. Le principal inconvénient de cette méthode, c'est que le système propose déjà une interprétation du guide à l'expert : cela peut biaiser son jugement.

La seconde approche consiste à confronter l'annotation obtenue par l'outil à celle obtenue par l'expert qui opère manuellement. Par cette approche, l'expert peut formuler sa propre interprétation du guide, cependant ce travail est très fastidieux.

Les GBPC sont souvent complexes et d'interprétation difficile, le recours à plusieurs experts risque d'aboutir a plusieurs interprétations différentes. Il est possible de mesurer le niveau d'accord entre les experts par le coefficient du Kappa mais cela n'indique pas quelle est l'interprétation juste et si l'outil s'en approche. Pour s'affranchir des variations dans l'interprétation des guides, il est possible de chercher le consensus en utilisant la méthode Delphi : les désaccords entre experts sont relevés, et on redemande au experts de réinterpréter le guide à la lumière des interprétations des autres experts, et ceci jusqu'à ce que l'ensemble des experts se mettent d'accord pour produire une référence (*gold standard*).

La confrontation des résultats de l'outil à une référence (*gold standard*) peut s'exprimer à l'aide des indicateurs de rappel (ratio d'information correcte extraite sur l'information disponible) et de précision (ratio de l'information correcte sur la totalité de l'information extraite) . Le système doit privilégier une mesure de rappel élevé (et idéalement égale à 100%) et une mesure de précision réduite car il est plus facile d'enlever de l'information non pertinente que de réinterpréter la totalité d'un guide en raison d'un défaut de détection d'information pertinente.

### 4.2 Mise en oeuvre de l'évaluation de l'outil GemFrame

Nous présentons dans cette partie quelques éléments d'évaluation du travail effectué sur les guides de bonnes pratiques.

#### 4.2.1 Élaboration de la référence manuelle

Un certain nombre de GBPC ont été annotés manuellement pour servir de base à l'étude et pour servir lors de l'évaluation. L'évaluation est faite sur des GBPC n'ayant pas servi lors de la mise au point du système.

Globalement, chaque annotateur doit dériver à partir d'un GBPC un arbre XML conforme à la notation GEM. L'accord entre annotateurs est ensuite calculé en comparant le nombre de recommandations reliées à l'action ou à l'ensemble d'actions dont elle dépend.

L'accord entre annotateurs, calculé à partir d'un ensemble de 162 nœuds, est de 0,96 (157 segments correctement raccordés sur 162). Ces résultats montrent la faisabilité de la tâche et une variation dans l'annotation réduite, au moins sur certains GBPC. Des expériences faites en comparant une annotation non-experte avec une annotation experte note seulement des différences minimes. Ceci permet de dégager deux conséquences : 1) l'annotation ne repose que légèrement sur des connaissances expertes et 2) elle peut être faite manuellement en s'appuyant sur les caractéristiques linguistiques et typo-dispositionnelles, dans la mesure où les connaissances du domaine ne sont que peu sollicitées.

D'autres études ont mis en évidence une plus grande variabilité entre annotateurs humains (cf. *supra*, introduction). Il serait intéressant de prolonger l'expérience pour voir dans quelle mesure ces résultats varient en fonction du GBPC analysé, des guides d'annotations fournis et des annotateurs concernés.

### 4.2.2 Évaluation du découpage en segments élémentaires

On évalue le découpage en segments élémentaires en calculant des scores de rappel et de précision. On obtient alors les résultats suivants (la F-mesure est la moyenne harmonique du rappel et de la précision).

| **GBP : Cancer du sein** | | | |
|---|---|---|---|
| | **Conditions** | **Recommandations** | **Rattachements** |
| # présents | 73 | 96 | 169 |
| # trouvés | 60 | 88 | |
| *rappel* | *82,19* | *91,66* | |
| # corrects | 70 | 94 | 126 |
| *précision* | *95,89* | *97,91* | ***74,55*** |
| **F-mesure** | **88,85** | **94,64** | |

| **GBPC : Critères chimio** | | | |
|---|---|---|---|
| | **Conditions** | **Recommandations** | **Rattachements** |
| # présents | 70 | 107 | 177 |
| # retrouvés | 61 | 96 | |
| *rappel* | *87,14* | *89,71* | |
| # corrects | 65 | 104 | 136 |
| *précision* | *92,85* | *97,19* | ***76,83*** |
| **F-mesure** | **90,05** | **93,29** | |

| **GBPC : Dénutrition des personnes âgées** | | | |
|---|---|---|---|
| | **Conditions** | **Recommandations** | **Rattachements** |
| # présents | 75 | 107 | 182 |
| # retrouvés | 62 | 100 | |
| *rappel* | *82,66* | *93,45* | |
| # corrects | 73 | 106 | 162 |
| *précision* | *97,33* | *99,06* | ***89,01*** |
| **F-mesure** | **89,34** | **95,65** | |

| **GBPC : AOMI** | | | |
|---|---|---|---|
| | **Conditions** | **Recommandations** | **Rattachements** |
| # présents | 60 | 91 | 151 |
| # retrouvés | 45 | 65 | |
| *rappel* | 75 | 71,42 | |
| # corrects | 59 | 88 | 107 |
| *précision* | 98,33 | 96,70 | **70,86** |
| **F-mesure** | **85,08** | **82,14** | |

Pour les rattachements des recommandations aux conditions (dernière colonne), comme on compare le nombre d'éléments correctement rattachés par rapport à la référence, on obtient juste un score de précision (*accuracy*).

**Evaluation du découpage en segments élémentaires**

Les résultats obtenus sont relativement bons, aussi bien pour la reconnaissance des recommandations que pour la reconnaissance des conditions. L'apport des connaissances du domaine n'est pas évident à la vue des résultats. Cette information est cependant pertinente quand il s'agit d'étiqueter des éléments isolés, par exemple des titres correspondant à des pathologies. Ainsi, le titre *hypertension artérielle* est équivalent à une condition introduite par *en cas de*. Il est donc important de reconnaître ces cas et de les analyser correctement dans la mesure où plusieurs recommandations sont susceptibles d'être sous la portée de cette condition. Cette analyse ne peut être faite sans connaissances du domaine.

Le nombre et la nature des titres diffèrent profondément d'un guide à l'autre. Quand le nombre de titres de ce type est important, l'impact des connaissances du domaine n'est pas négligeable dans la mesure où la non-analyse des titres peut faire échouer le bon rattachement toute une série de recommandations.

Enfin, on peut constater que tous les titres n'ont pas la même importance (certains sont des noeuds centraux, ils contrôlent davantage de recommandations, *etc.*) mais il est difficile de tenir compte de cet aspect dans l'évaluation.

### 4.2.3 Évaluation de la portée des conditions

La portée des conditions est analysée correctement dans 77 % des cas (ce chiffre est obtenu en extrayant, à partir de l'annotation manuelle, l'ensemble des couples simples condition/recommandation. On compare ensuite le nombre de couples communs entre la référence obtenue par l'étiquetage manuel et les données produites par le système. On obtient ainsi le taux de précision de l'analyseur).

Ce résultat est encourageant, surtout si on prend en compte l'ensemble des paramètres impliqués lors de l'analyse du discours. Dans la plupart des cas, la portée de la condition est obtenue à partir de l'application des règles par défaut. Cependant, quelques cas importants sont résolus grâce aux règles fondées sur l'analyse des exceptions, qui permettent d'aller en deçà ou au-delà de la portée par défaut.

Le système échoue particulièrement en cas de portée étendue, quand celle-ci est exprimée par des marques cohésion liées au vocabulaire du domaine (utilisation de synonymes, d'hyponymes ou d'hyperonyme) ou de structures complexes (anaphores nominales, structures syntaxiques complexes). Résoudre ces cas rares reviendrait à introduire de nombreuses connaissances du domaine dans le système, avec tous les risques que cela comporte. Nous avons volontairement écarté cette voie dans la mesure où on souhaite garder un système relativement portable, faisant un usage minimal des connaissances du domaine.

## 5   Conclusion

Nous avons présenté dans cet article un outil d'aide à la modélisation des GBPC. Ces documents nécessitent une modélisation, mais ce travail manuel est long, fastidieux et variable d'un individu à l'autre. L'automatisation de la tâche permet de s'affranchir partiellement de la variation individuelle en proposant une méthode fixe, définie et reproductible.

Les résultats obtenus sont satisfaisants eu égard à la difficulté de la tâche. Une étude plus précise des cas d'échecs serait nécessaire afin d'affiner l'outil. Il faut toutefois noter que certains cas semblent difficilement automatisables; surtout si l'on veut garder une certaine généricité dans les traitements. L'outil est donc une aide à la modélisation ; il ne prétend en aucun cas remplacer l'analyste humain. Il doit au contraire permettre à celui-ci de se concentrer sur les cas difficiles en automatisant le traitement des cas les cas simples.

## Références